\documentclass[sigconf,screen,nonacm,10pt]{acmart}

\AtBeginDocument{%
}

\setcopyright{none}
\acmConference{}{}{}
\acmBooktitle{}
\acmDOI{}
\acmISBN{}
\authorsaddresses{}

\usepackage{graphicx}
\usepackage{amsmath}

\usepackage{amssymb}
\usepackage{booktabs}
\usepackage{microtype}
\usepackage{algorithm}
\usepackage{algpseudocode}
\usepackage{xcolor}
\definecolor{sinkteal}{RGB}{24,104,96}

\fancypagestyle{arxiv}{%
  \fancyhf{}%
  \fancyfoot[C]{\footnotesize\thepage}%
}

\newcommand{\safeincludegraphics}[2][]{%
  \IfFileExists{#2}{%
    \includegraphics[#1]{#2}%
  }{%
    \PackageWarning{ssha}{Missing figure file: \detokenize{#2}}%
    \fbox{%
      \parbox[c][0.18\textheight][c]{0.92\linewidth}{%
        \centering Missing figure file: \texttt{\detokenize{#2}}%
      }%
    }%
  }%
}

\newcommand{\vbenchavg}{VBench Avg (8 dims)}
\newcommand{\softtifagm}{Soft-TIFA GM}
\newcommand{\softtifaam}{Soft-TIFA AM}

\begin{document}

\title[Sink-Sparse Hybrid Attention]{Sink-Sparse Hybrid Attention for Efficient Inference in T2I/T2V Diffusion Transformers}

\author{Bingxin Xing}
\authornote{Equal contribution.}
\affiliation{%
  \institution{Xi'an Jiaotong University}
  \city{Xi'an}
  \country{China}
}

\author{Yu Zhang}
\authornotemark[1]
\affiliation{%
  \institution{Xiamen University}
  \city{Xiamen}
  \country{China}
}

\author{Jiatong Ding}
\affiliation{%
  \institution{Shanghai Jiao Tong University}
  \city{Shanghai}
  \country{China}
}

\author{Lei Lu}
\affiliation{%
  \institution{Xi'an Jiaotong University}
  \city{Xi'an}
  \country{China}
}

\author{Ping Wang}
\authornote{Corresponding author.}
\affiliation{%
  \institution{Xi'an Jiaotong University}
  \city{Xi'an}
  \country{China}
}

\renewcommand{\shortauthors}{Bingxin Xing et al.}

\newcommand{\SSHAauthorblock}{%
  \begingroup
  \centering
  {\normalfont\rmfamily\fontsize{11}{13}\selectfont\bfseries
    Bingxin Xing\textsuperscript{\normalfont *1},\quad
    Yu Zhang\textsuperscript{\normalfont *2},\quad
    Jiatong Ding\textsuperscript{\normalfont 3},\quad
    Lei Lu\textsuperscript{\normalfont 1},\quad
    Ping Wang\textsuperscript{\normalfont \textdagger 1}\par
  }%
  \vspace{5pt}%
  {\normalfont\rmfamily\normalsize
    \textsuperscript{1}Xi'an Jiaotong University\quad
    \textsuperscript{2}Xiamen University\quad
    \textsuperscript{3}Shanghai Jiao Tong University\par
  }%
  \endgroup
}
\makeatletter
\def\@mkauthors@iii{%
  \gdef\@currentauthors{}%
  \gdef\@currentaffiliation{}%
  \global\setbox\mktitle@bx=\vbox{%
    \hsize=\textwidth
    \unvbox\mktitle@bx
    \SSHAauthorblock
    \par\bigskip
  }%
}
\makeatother

\begin{abstract}
Diffusion Transformers (DiTs) have become a common backbone for high-quality text-to-image and text-to-video generation, but repeated dense self-attention remains a major inference bottleneck as token sequences grow and the same operator is invoked across many denoising steps. Our study suggests that dense attention in pretrained DiTs contains reusable layer/head-specific high-mass structure that can be identified from a small offline calibration set and reused across future generations. This leads to Sink-Sparse Hybrid Attention, a train-free attention replacement that reuses static sink masks and supplements them with a lightweight pooled-KV branch, while leaving model weights and sampling schedules unchanged. On Wan2.1-T2V-1.3B, the method attains 1.54$\times$ end-to-end speedup while maintaining generation quality. On Z-Image-Turbo, it attains 1.57$\times$ end-to-end speedup with competitive GenEval2 performance. These results indicate that reusable importance-aware sparse support can provide an effective train-free acceleration path for both T2V and T2I DiTs.
\end{abstract}


\maketitle
\pagestyle{arxiv}
\thispagestyle{arxiv}

\hypersetup{%
  pdftitle={Sink-Sparse Hybrid Attention for Efficient Inference in T2I/T2V Diffusion Transformers},
  pdfauthor={Bingxin Xing, Yu Zhang, Jiatong Ding, Lei Lu, Ping Wang},
  pdfsubject={Train-free sparse attention for diffusion transformers},
  pdfkeywords={diffusion transformers, sparse attention, train-free acceleration, text-to-image, text-to-video},
  pdfpublisher={}
}

\section{Introduction}

Diffusion Transformers (DiTs) have become an important backbone for high-quality visual generation in both text-to-image (T2I) and text-to-video (T2V) systems~\cite{peebles2023dit,bao2023uvit,polyak2025moviegen,wan2025wan,zimage2025zimage}. As these systems move toward higher resolutions, longer outputs, and repeated sampling in practical workflows, inference efficiency becomes a first-order deployment concern. The relevant cost is not only the latency of a single sample: users often regenerate multiple candidates under different prompts, seeds, or motion descriptions, and online services must support many such requests concurrently. Cumulative latency and hardware budget therefore become dominant practical constraints. This pressure is especially pronounced in modern visual generation because quality improvements often come from exactly the factors that increase compute demand: more denoising steps, larger latent grids, longer videos, and richer global token interaction.

A major source of this cost is self-attention. In video DiTs, self-attention must model all interactions over long spatio-temporal token sequences, and this computation is repeated across layers, heads, and denoising steps~\cite{vaswani2017attention,liu2025sta,xia2025adaspa}. In high-resolution image DiTs, the temporal axis disappears, but the large spatial token grid still makes repeated global interaction expensive. Although optimized dense kernels can substantially reduce implementation overhead, they do not change the underlying quadratic interaction pattern~\cite{dao2022flashattention}. Consequently, attention remains a major runtime bottleneck even when the rest of the inference stack is already carefully optimized. Because diffusion sampling evaluates nearby latent states over many denoising iterations, the same expensive operator is invoked again and again; any inefficiency is therefore multiplied along the full trajectory. As later shown in Sec.~\ref{sec:motivation_cost}, self-attention alone accounts for a large fraction of end-to-end sampling time in the evaluated setting.

This bottleneck has motivated extensive research on efficient diffusion inference. Some methods reduce redundant computation across denoising steps through feature caching or cross-step reuse, such as DeepCache, TGATE, FasterCache, ToCa, and AB-Cache~\cite{ma2024deepcache,liu2024tgate,lv2024fastercache,zou2025toca,yu2025abcache}. Another line targets the attention operator itself. For video DiTs, train-free methods such as SVG, SVG2, STA, and AdaSpa exploit spatial-temporal sparsity, local windows, or adaptive sparse search to accelerate pretrained backbones~\cite{svg1,svg2,liu2025sta,xia2025adaspa}. In parallel, more generic sparse operators such as XAttention and SpargeAttn provide reusable block selection and sparse execution mechanisms that can also be instantiated in visual generation models~\cite{xu2025xattention,zhang2025spargeattention}. Training-based methods such as VSA and Video-BLADE instead learn sparse behavior through optimization or distillation~\cite{zhang2025vsa,gu2025blade}. Together, these studies show that dense attention is not the only viable operating point. Yet for deployment, the most attractive solution is one whose sparse structure can be prepared once, reused across diverse generation requests, and integrated into existing inference stacks without retraining, fine-tuning, or architecture-specific redesign.

However, train-free sparse inference still faces a persistent difficulty. When the pretrained backbone is frozen, the outcome depends not only on \emph{how much} computation is removed, but also on \emph{which} interactions are preserved. Existing sparse rules often rely on locality priors, online search, or generic scoring functions. These choices can be effective, yet they do not directly answer a more reusable question: can a small subset of important interactions be identified once offline and then amortized across future generations? Local windows are simple and hardware-friendly, but long-range semantic or temporal dependencies do not always respect a fixed neighborhood. More adaptive selection rules can recover missing interactions, but they also introduce online overhead and repeated pattern discovery. In other words, train-free sparse inference often has to choose between cheap but rigid sparse layouts and flexible but repeatedly re-computed sparse layouts.

Prior work already suggests that attention in diffusion transformers is structured rather than arbitrary. Locality-based methods reveal strong spatial-temporal regularity, while adaptive methods exploit cross-step or input-dependent sparsity~\cite{svg1,svg2,liu2025sta,xia2025adaspa}. Our question is more specific: can a small offline calibration set identify layer/head-specific block interactions that remain useful across prompts, noise seeds, and denoising steps, so that a static mask can be reused with low online overhead? If the answer is yes, train-free sparse attention need not choose only between rigid hand-designed layouts and repeated online pattern discovery.

This question also aligns with importance-aware findings in long-context transformers. Attention-sink, heavy-hitter, KV-selection, and structured sparse-prefill studies show that a relatively small subset of tokens or regions can repeatedly dominate attention and stabilize compressed inference~\cite{xiao2023streamingllm,zhang2023h2o,xiao2024duoattention,li2024snapkv,jiang2024minference,lai2025flexprefill,zhang2026s2o}. Although diffusion attention differs substantially from causal LLM decoding, these results support the broader intuition that preserving recurrent high-importance structure can be more effective than treating all interactions as equally necessary. For diffusion transformers, the relevant reusable unit is not a causal token sink but a layer/head-specific block-level support pattern in bidirectional attention.

Our key observation is that a small calibration set can expose reusable block-level supports that repeatedly capture a large fraction of dense attention mass across prompts, noise seeds, and denoising steps. We refer to these retained regions as \emph{sink blocks}. Unlike attention-sink tokens in causal LLMs, sink blocks here are layer/head-specific block supports in bidirectional attention. This observation makes static sparsification attractive, but it also exposes its limitation: the dominant reusable support captures most, not all, of the useful context. Discarding the remaining support entirely can remove weak but globally distributed interactions, as well as input-dependent context that is less stable than the main sink structure. This is why we use a hybrid attention design rather than a sink-only sparse operator.

We instantiate this idea in \textbf{Sink-Sparse Hybrid Attention}, a structure-aware, train-free replacement for dense self-attention in pretrained DiTs. For each layer and head, a static sink mask is calibrated once offline and then reused during inference. A sink-preserving sparse branch handles the recurrent high-mass interactions that can be amortized across runs, while a pooled-KV compensation branch provides a low-cost approximation to the residual global context using full-resolution queries and pooled keys/values. The two branches are fused with a normalization-aware rule so that the sparse sink branch remains dominant when its retained support is sufficient, while the compensation branch contributes when omitted context becomes important. The resulting workflow is simple and deployment-friendly, requiring no retraining, fine-tuning, or sampling-schedule changes.

Our contributions are summarized as follows:
\begin{itemize}
    \item We show that a small calibration set can identify reusable block-level static masks that remain well aligned with dense attention across prompts, noise seeds, and denoising steps.
    \item We develop \textbf{Sink-Sparse Hybrid Attention}, a train-free attention replacement that combines per-layer, per-head static sink masks with a pooled-KV compensation branch and normalization-aware fusion, and we clarify why the hybrid branch is useful for recovering coarse omitted context at low cost.
    \item We validate the framework on both T2V and T2I diffusion transformers, showing favorable quality--efficiency trade-offs across the two generation settings and demonstrating that hybrid compensation improves the practicality of reusable train-free sparsity.
\end{itemize}

\section{Related Work}

\subsection{Efficient Inference for Diffusion Models}

A broad literature studies how to reduce the cost of diffusion sampling. One important direction, especially in train-free settings, exploits redundancy across denoising steps through feature caching, attention reuse, or selective recomputation. Representative examples include DeepCache, TGATE, FasterCache, ToCa, and AB-Cache~\cite{ma2024deepcache,liu2024tgate,lv2024fastercache,zou2025toca,yu2025abcache}. These methods are effective because diffusion sampling repeatedly evaluates nearby latent states, so part of the computation can often be reused rather than recomputed from scratch.

Another direction improves diffusion efficiency through attention approximation, architectural redesign, or task-specific acceleration. DiTFastAttn targets faster attention in pretrained diffusion backbones, whereas DiG redesigns the generator itself, and recent editing-oriented methods accelerate related diffusion workflows~\cite{yuan2024ditfastattn,pu2024eedit,zhu2025dig,becker2025edit}. Together, these studies show that attention is not merely a low-level kernel concern but a major algorithmic target for practical diffusion inference acceleration.

\subsection{Sparse Attention in Diffusion Transformers}

Sparse attention is especially attractive for DiTs because transformer-based generators inherit the high cost of dense token interaction while evaluating the same operator many times during sampling. In the T2V setting, train-free methods such as SVG, SVG2, STA, and AdaSpa exploit spatial-temporal locality, sparse support, or adaptive search to accelerate pretrained video DiTs~\cite{svg1,svg2,liu2025sta,xia2025adaspa}. These methods demonstrate that video-DiT attention is far from unstructured and that substantial end-to-end savings are possible without changing model weights.

In parallel, more operator-oriented approaches such as SpargeAttn and XAttention emphasize reusable sparse selection rules and efficient sparse execution primitives that can transfer across workloads~\cite{zhang2025spargeattention,xu2025xattention}. A separate line of work, including VSA and Video-BLADE, incorporates sparsity through training, adaptation, or distillation~\cite{zhang2025vsa,gu2025blade}. Compared with these methods, our goal is deliberately narrow and practical: we focus on the train-free setting and ask whether a small offline calibration stage can identify a reusable sparse support that is cheap to deploy later.

\subsection{Importance-Aware and Structured Sparsity in Transformers}

Our work is also related to a broader importance-aware view of transformer sparsity. In long-context language models, attention-sink, heavy-hitter, KV-selection, and structured sparse-prefill methods show that a relatively small subset of tokens or regions can repeatedly dominate attention and support efficient approximate inference~\cite{xiao2023streamingllm,zhang2023h2o,xiao2024duoattention,li2024snapkv,jiang2024minference,lai2025flexprefill,zhang2025anchorattention,wang2025pbsattn,zhang2026s2o}. Although these methods were developed primarily for causal language modeling or prefilling rather than bidirectional diffusion attention, they provide an important conceptual cue: preserving recurrently important structure can be more effective than treating all interactions as equally necessary.

Our method adopts this perspective in a different regime. Instead of causal token sinks or online heavy-hitter selection, we study layer/head-specific block-level support in pretrained DiTs and reuse it across denoising steps and future runs. In this sense, Sink-Sparse Hybrid Attention occupies a middle ground between fully hand-designed sparse layouts and fully online pattern discovery: the sparse structure is empirical and model-specific, but most of its cost is paid once during offline calibration.

\section{Motivation}
\label{sec:motivation}

This section asks two questions: (1) whether self-attention is a major runtime bottleneck, and (2) whether dense attention exhibits reusable structure that can support a static sparse pattern.

Beyond raw speed, we are interested in whether attention structure is stable enough to justify paying the pattern-discovery cost only once. A useful train-free sparse operator should therefore satisfy two practical criteria: it should target a component that occupies a substantial portion of end-to-end latency, and the retained support should remain informative across prompts, seeds, and denoising phases. The following evidence is organized around exactly these two requirements and motivates why a reusable static mask is a reasonable design target.

\subsection{Self-Attention Is a Major Runtime Bottleneck}
\label{sec:motivation_cost}

\begin{figure}[t]
    \centering
    \safeincludegraphics[width=0.92\linewidth]{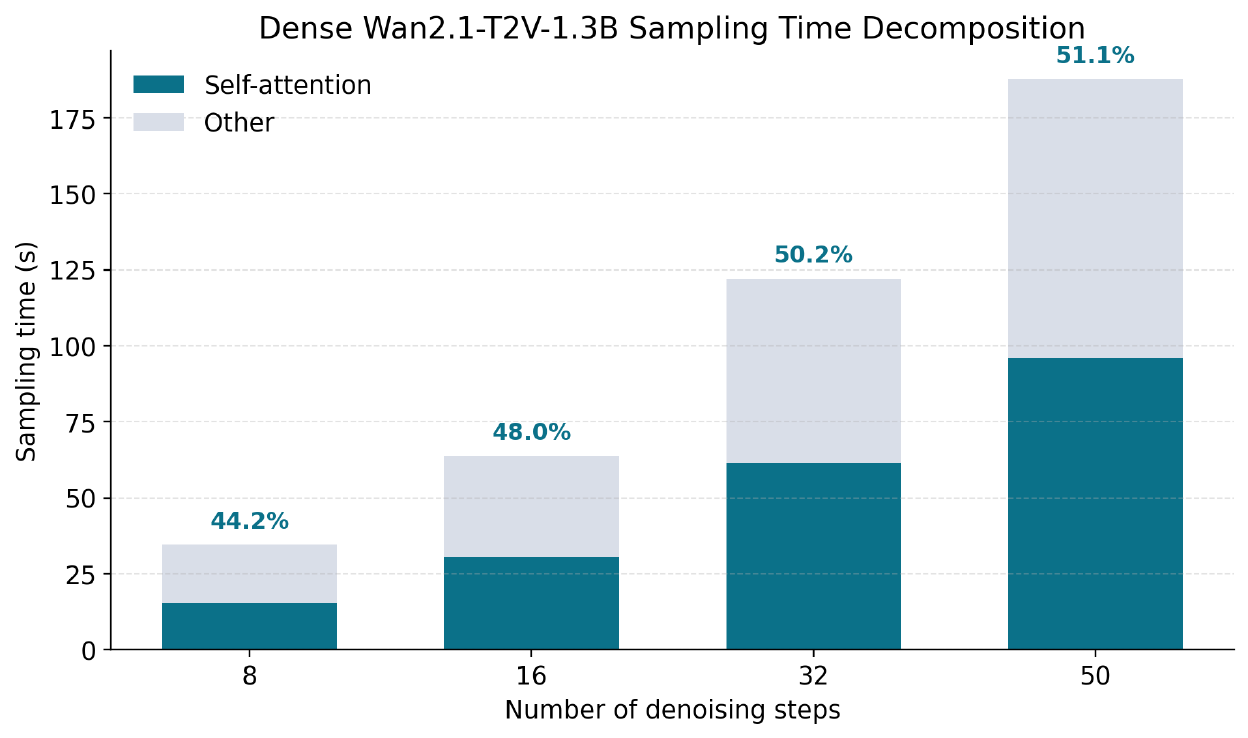}
    \caption{\textbf{Profiling sampling latency in the evaluated T2V setting.} We decompose the sampling time $T_{\mathrm{sample}}$ (excluding model initialization and video writing) into self-attention time $T_{\mathrm{attn}}$ and the remaining components $T_{\mathrm{other}}$ across different denoising schedules. Self-attention consistently occupies a large fraction of the total runtime.}
    \Description{Bar chart profiling sampling latency across denoising schedules, separating self-attention time from the remaining runtime and showing that attention occupies a large fraction of inference cost.}
    \label{fig:motivation_cost}
\end{figure}

We first verify that self-attention is a meaningful target for acceleration. Under the evaluated long-sequence setting, self-attention consistently accounts for about \textbf{44\%--51\%} of sampling time across denoising schedules, as shown in Fig.~\ref{fig:motivation_cost}. This confirms that reducing attention cost can translate into end-to-end savings rather than only micro-kernel improvements.

The profiling result also clarifies why a train-free attention replacement is worth studying. The same high-cost attention operator is invoked repeatedly along the denoising trajectory, so even a moderate per-step reduction can accumulate into substantial wall-clock savings. At the same time, the success of prior sparse-attention methods suggests that dense attention contains structured redundancy: some interactions appear much more important than others and therefore should be preserved more carefully.

\begin{figure}[t]
    \centering
    {\footnotesize \textbf{Wan2.1-T2V-1.3B}}\\[2pt]
    \safeincludegraphics[width=0.95\linewidth]{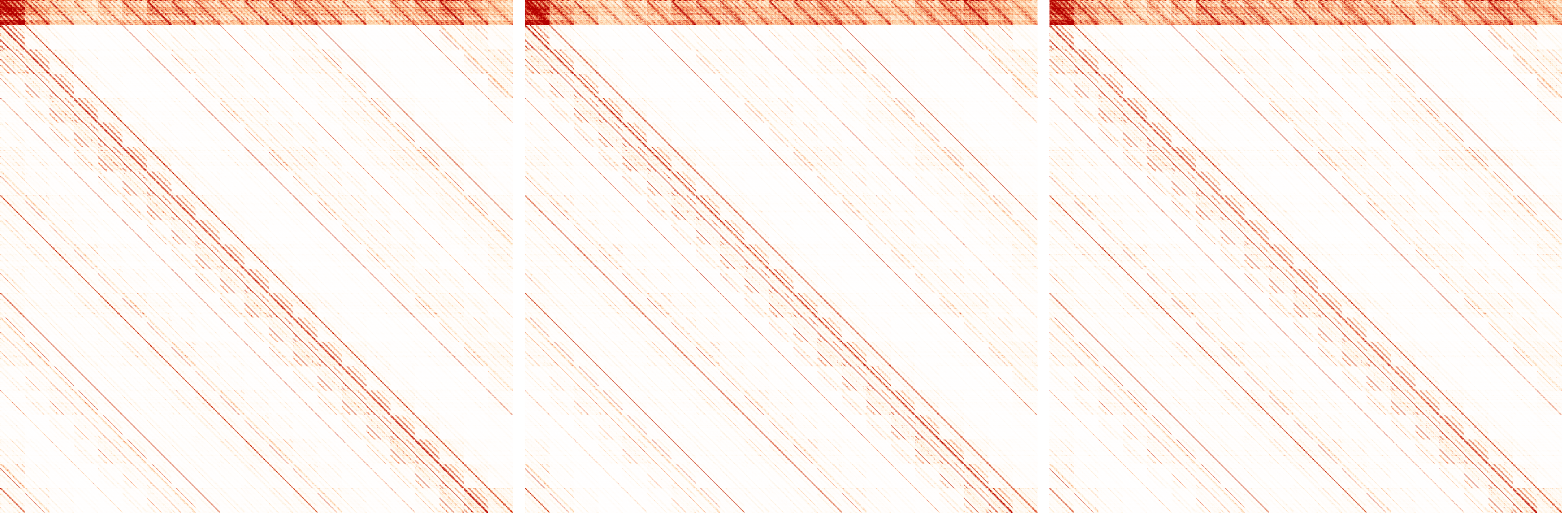}\\[2pt]
    {\footnotesize \textbf{Z-Image-Turbo}}\\[2pt]
    \safeincludegraphics[width=0.95\linewidth]{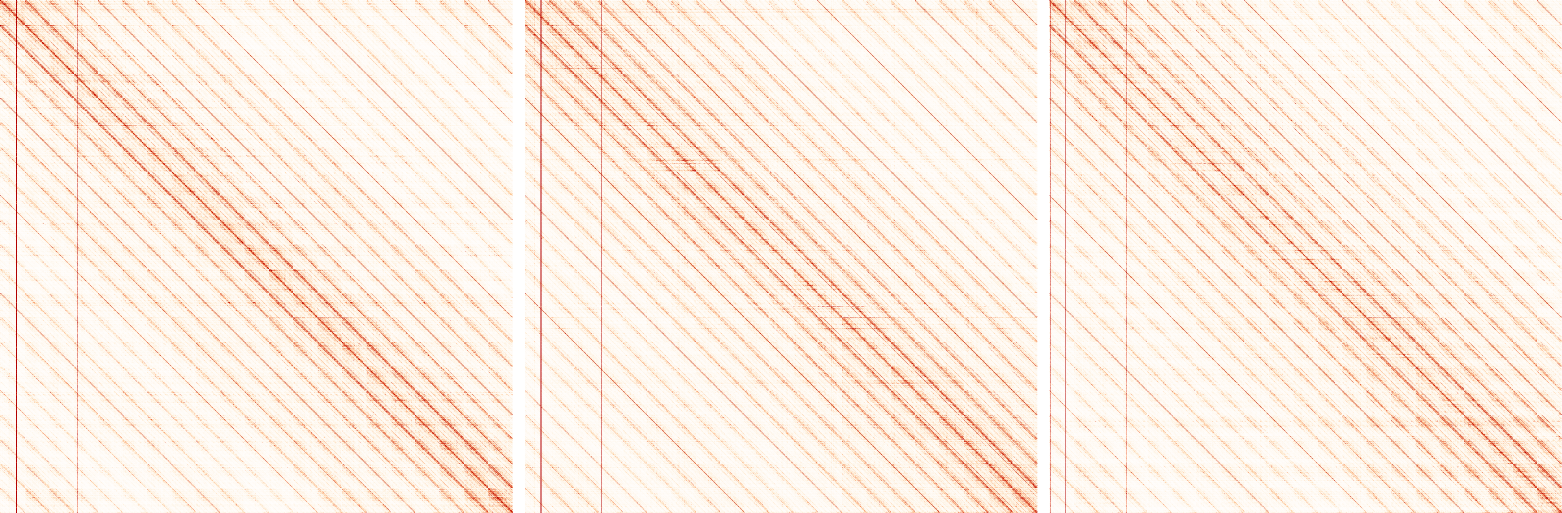}
    \caption{\textbf{Attention pattern similarity across different prompts.} We visualize representative dense self-attention maps from \textbf{Wan2.1-T2V-1.3B} and \textbf{Z-Image-Turbo}, each under three different prompts. Despite changes in prompt content, the dominant stripe-like structures remain visually similar.}
    \Description{Representative dense self-attention heatmaps from Wan2.1-T2V-1.3B and Z-Image-Turbo, each showing three different prompts with visually similar stripe-like structures.}
    \label{fig:motivation_similarity}
\end{figure}

\subsection{Stable High-Mass Attention Regions Across Inputs and Steps}
\label{sec:motivation_similarity}

We next ask whether dense attention exhibits a reusable structure that could support a static sparse pattern. When the same layer and head are visualized under different prompt--seed pairs, prominent stripe-like bands and recurrent long-range links remain visible even though local intensities change, as shown in Fig.~\ref{fig:motivation_similarity}. This indicates that the attention pattern contains a persistent structural component in addition to instance-specific variation. Our goal is not to claim that the dense maps are identical across runs, but rather that a stable block-level scaffold persists strongly enough to guide a reusable sparse operator.

\begin{figure}[t]
    \centering
    \safeincludegraphics[width=0.95\linewidth]{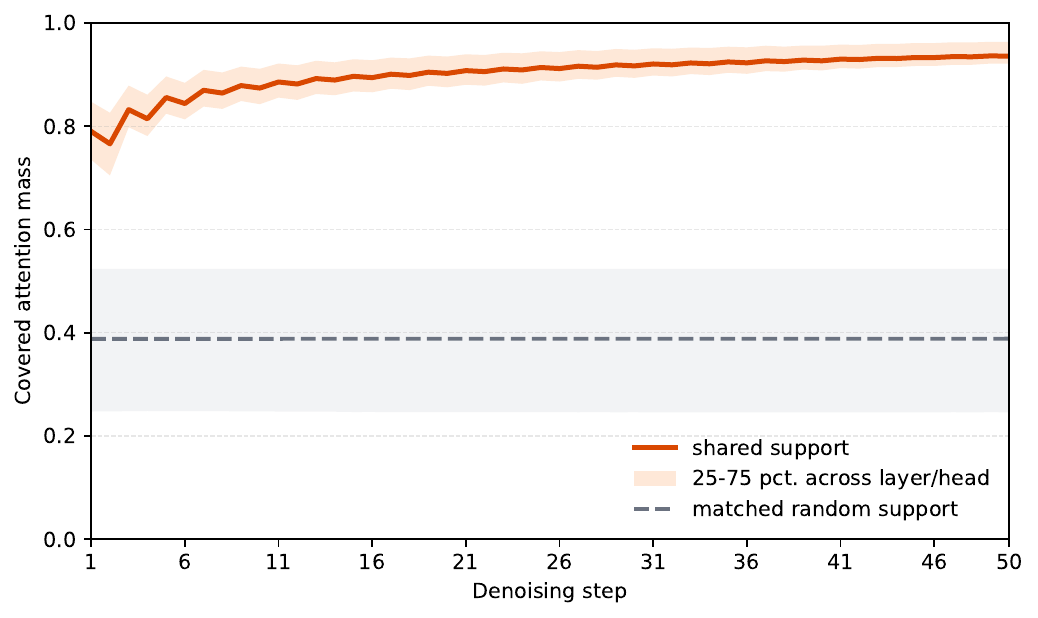}
    \caption{\textbf{Coverage of dense attention mass captured by the static sink mask across denoising steps.} We evaluate a representative shared-step static sink mask with mean retained block density $38.8\%$ (i.e., $61.2\%$ sparsity). For each denoising step, the sink-mask curve shows the mean coverage across layer--head pairs on held-out cases, and the shaded band shows the 25th--75th percentile range. A budget-matched random mask is shown as a reference.}
    \Description{Coverage-versus-step plot for a representative shared-step static sink mask, showing the fraction of dense attention probability mass preserved throughout the default 50-step denoising schedule, together with a budget-matched random baseline and shaded 25th--75th percentile ranges across layer--head pairs.}
    \label{fig:motivation_step_mask}
\end{figure}

To quantify this behavior at the model level, we measure how much dense attention probability mass falls inside the token-level expansion $\widetilde{M}^{(l,h)}$ of a layer/head-specific static sink mask. For a held-out prompt--seed case $u$, denoising step $t$, layer $l$, and head $h$, let $P_{u}^{(l,h,t)}=\operatorname{softmax}(S_{u}^{(l,h,t)})$ denote the dense attention probability matrix. We define the corresponding coverage as
\begin{equation}
\mathrm{Cov}_{u,l,h,t}
=
\frac{\sum_{i,j} [P_{u}^{(l,h,t)}]_{ij}\,\widetilde{M}^{(l,h)}_{ij}}
{\sum_{i,j} [P_{u}^{(l,h,t)}]_{ij}}.
\label{eq:motivation_cov_single}
\end{equation}
We report this quantity in Fig.~\ref{fig:motivation_step_mask} for a representative shared-step static sink mask whose mean retained block density is about $38.8\%$ (equivalently, about $61.2\%$ sparsity). For each denoising step, we summarize the statistic over held-out cases and all layer--head pairs by the mean and the 25th--75th percentile band. A budget-matched random mask is shown only as a reference.

Figure~\ref{fig:motivation_step_mask} shows that the static sink mask retains a high fraction of dense attention mass throughout the default 50-step denoising trajectory. Coverage is already high at early steps, shows some fluctuation in the initial phase, and then remains high throughout the rest of denoising. The reference baseline remains much lower, but the more important point is that the same reusable mask stays well aligned with dense attention over a wide range of inputs and steps.

\paragraph{Implication.}
These observations make a static train-free design plausible. If recurrent high-mass regions persist across runs, they can be preserved by a reusable sink mask. At the same time, coverage is clearly below one, so a pure sink-only approximation would still discard some weaker but useful context. This motivates a hybrid design: the sink branch preserves the stable scaffold, while a lightweight complementary branch restores coarse global information omitted by the static mask.

\section{Method}
\label{sec:method}

\begin{figure*}[t]
    \centering
    \safeincludegraphics[width=0.98\textwidth]{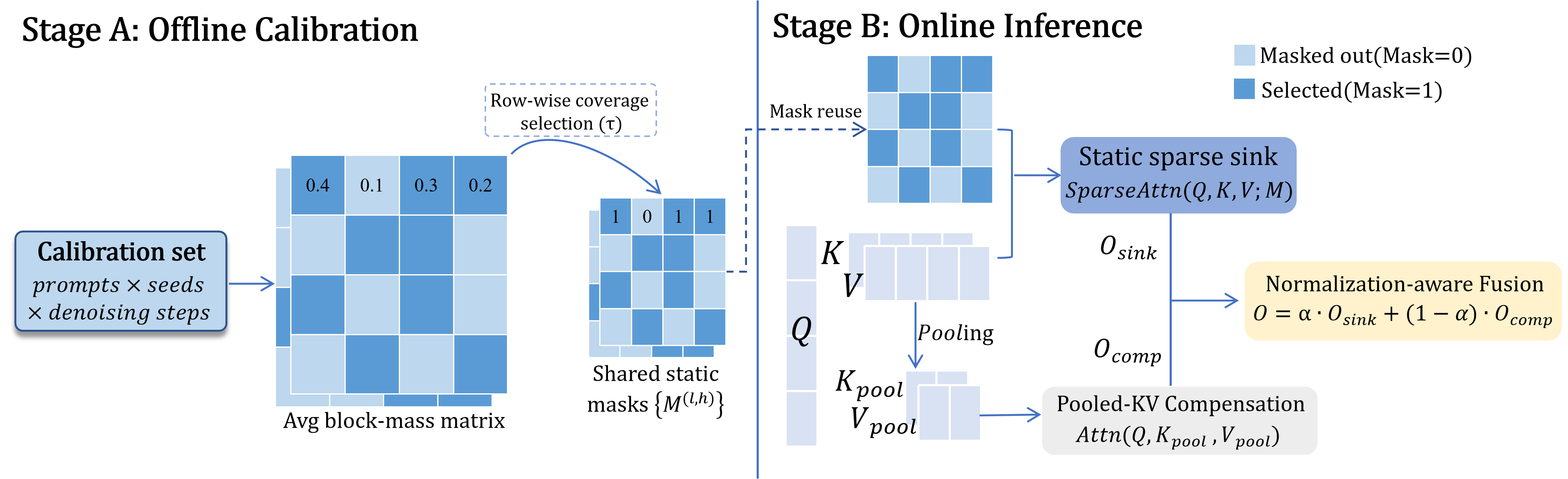}
    \caption{\textbf{Workflow of Sink-Sparse Hybrid Attention.}
    \textbf{Offline calibration} collects dense self-attention observations from a small calibration set, converts token-level attention into exact block-mass scores, averages them across observations, and applies a row-wise coverage threshold $\tau$ to obtain one static sink mask for each layer and head.
    \textbf{Online inference} reuses these masks at every denoising step: a static sparse sink branch preserves recurrently important interactions, a pooled-KV compensation branch supplies coarse omitted context using full-resolution queries and pooled keys/values, and the two branches are fused into the final attention output.}
    \Description{Two-stage workflow of Sink-Sparse Hybrid Attention. The offline stage measures dense token attention, aggregates it into exact block-level mass, averages it across calibration observations, and constructs one static sink mask per layer and head through row-wise coverage selection. The online stage reuses the same static sink masks at every denoising step, combines a static sparse sink branch with a pooled-KV compensation branch, and fuses their outputs into the final attention result.}
    \label{fig:method_workflow}
\end{figure*}

\subsection{Notation and Overview}
\label{sec:method_notation}

For layer $l$ and head $h$, dense self-attention is
\begin{equation}
O^{(l,h)} =
\operatorname{softmax}\!\left(
\frac{Q^{(l,h)}(K^{(l,h)})^\top}{\sqrt{d_h}}
\right)V^{(l,h)}.
\label{eq:method_dense_attn_compact}
\end{equation}
Let the flattened token sequence contain $N$ tokens. We partition it into $B$ consecutive blocks of size $b$ (assuming $N$ is divisible by $b$ for simplicity), and estimate one static sink mask
$M^{(l,h)} \in \{0,1\}^{B \times B}$ for each layer and head.
In this paper, we use the original token order; no rearrangement is applied before mask construction.
The same static sink mask is reused at every denoising step.
For different backbones, the masks are calibrated independently.

The full pipeline has two stages.
First, we build static sink masks from a small dense calibration set.
Second, we reuse these masks during inference in either a sink-only sparse branch or a hybrid branch with pooled-KV compensation.

\begin{algorithm}[t]
\caption{Offline Static Sink Mask Calibration}
\label{alg:offline_sink_mask}
\footnotesize
\begin{algorithmic}[1]
\Require calibration set $\mathcal{C}$, block size $b$, coverage threshold $\tau$
\Ensure static sink masks $\{M^{(l,h)}\}$
\Statex \textcolor{sinkteal}{\# initialize block-mass accumulators}
\ForAll{layers $l$ and heads $h$}
    \State $S^{(l,h)} \gets \mathbf{0}_{B \times B}$
\EndFor
\Statex \textcolor{sinkteal}{\# collect exact dense block-mass statistics}
\ForAll{observations $u \in \mathcal{C}$}
    \ForAll{layers $l$ and heads $h$}
        \State $A_u^{(l,h)} \gets \operatorname{softmax}\!\big(Q_u^{(l,h)}(K_u^{(l,h)})^\top / \sqrt{d_h}\big)$
        \State split the token sequence into consecutive blocks $\{\mathcal{B}_r\}_{r=1}^{B}$ of size $b$
        \State $m_u^{(l,h)} \gets \mathbf{0}_{B \times B}$
        \For{$r = 1$ to $B$}
            \For{$c = 1$ to $B$}
                \State $[m_u^{(l,h)}]_{r,c} \gets \frac{1}{|\mathcal{B}_r|}\sum_{i \in \mathcal{B}_r}\sum_{j \in \mathcal{B}_c} [A_u^{(l,h)}]_{ij}$
            \EndFor
        \EndFor
        \State $S^{(l,h)} \gets S^{(l,h)} + m_u^{(l,h)}$
    \EndFor
\EndFor
\Statex \textcolor{sinkteal}{\# build one static sink mask for each layer and head}
\ForAll{layers $l$ and heads $h$}
    \State $\bar{m}^{(l,h)} \gets S^{(l,h)} / |\mathcal{C}|$
    \State $M^{(l,h)} \gets \mathbf{0}_{B \times B}$
    \For{$r = 1$ to $B$}
        \State $\pi_r \gets \operatorname{argsort}\!\big(\bar{m}_{r,:}^{(l,h)}, \downarrow\big)$
        \State choose the shortest prefix $\pi_r[1{:}k_r]$ with $\sum_{t=1}^{k_r}\bar{m}_{r,\pi_r[t]}^{(l,h)} \ge \tau \sum_c \bar{m}_{r,c}^{(l,h)}$
        \State set $M_{r,\pi_r[1{:}k_r]}^{(l,h)} \gets 1$
    \EndFor
\EndFor
\State \Return $\{M^{(l,h)}\}$
\end{algorithmic}
\end{algorithm}

\subsection{Static Sink Mask Acquisition}
\label{sec:mask_acquisition}

Let $\mathcal{C}$ denote the set of calibration observations collected from several prompt--seed pairs and denoising steps.
No retraining or fine-tuning is involved.
The offline stage is designed to discover block pairs that repeatedly carry large dense attention mass across these observations.

\paragraph{Exact block-mass score.}
For a calibration observation $u \in \mathcal{C}$, let $\mathcal{B}_r$ and $\mathcal{B}_c$ denote the token sets of query block $r$ and key block $c$.
We first compute dense token-level attention probabilities and then aggregate them into block space:
\begin{equation}
\label{eq:method_block_mass}
\begin{aligned}
[A_u^{(l,h)}]_{ij}&=
\operatorname{softmax}_{j}\!\left(
\frac{Q_{i,u}^{(l,h)}(K_{j,u}^{(l,h)})^\top}{\sqrt{d_h}}
\right), \\
m_{r,c,u}^{(l,h)}
&=
\frac{1}{|\mathcal{B}_r|}
\sum_{i \in \mathcal{B}_r}
\sum_{j \in \mathcal{B}_c}
[A_u^{(l,h)}]_{ij}.
\end{aligned}
\end{equation}
This score is the exact block-level attention mass induced by dense attention. Because $\sum_c m_{r,c,u}^{(l,h)} = 1$ for every query-block row $r$, the coverage threshold $\tau$ has a direct interpretation: each row keeps the smallest set of key blocks whose averaged dense mass reaches at least $\tau$.
After averaging over the calibration set, we obtain
\begin{equation}
\bar{m}_{r,c}^{(l,h)}
=
\frac{1}{|\mathcal{C}|}
\sum_{u \in \mathcal{C}} m_{r,c,u}^{(l,h)}.
\label{eq:method_block_mass_avg}
\end{equation}

Algorithm~\ref{alg:offline_sink_mask} summarizes the offline calibration stage.
Its output is one static sink mask for each layer and head, which can be reused across denoising steps and future runs.

\subsection{Sink-Sparse Hybrid Inference}
\label{sec:train_free}

At inference time, the same static sink mask is reused at every denoising step.
The online module contains a static sparse sink branch and an optional pooled-KV compensation branch.
Using only the sink branch gives \textbf{Ours-Sink}; enabling both branches gives \textbf{Ours-Hybrid}.

\paragraph{Static sparse sink branch.}
The sink branch applies block-sparse self-attention only on the block pairs retained by $M^{(l,h)}$.
This branch preserves the recurrent high-mass interactions identified offline and provides the main computational savings.

\paragraph{Pooled-KV compensation branch.}
The hybrid variant adds a coarse complementary path.
The static sink mask is calibrated to preserve the recurrently dominant attention mass, but the omitted support is not completely negligible: some useful context is weaker, more distributed, or more input-dependent than the main sink structure.
Instead of recovering this missing support with another expensive dense branch, we use a low-cost approximation.
Queries remain at full resolution, while keys and values are compressed by non-overlapping mean pooling with gap $g_{\text{comp}}$.
This branch therefore serves as a coarse completion path that restores global context beyond the retained sink support.
Its usefulness is practical: sink-only sparsification preserves the dominant attention skeleton but can miss weak, distributed context.
The pooled branch summarizes this omitted global context at low cost and therefore stabilizes aggressive sparse settings.

\paragraph{Adaptive fusion.}
Because the two branches operate on different support sizes, their outputs should not be combined with a naive fixed average.
We therefore compute one row-wise fusion weight from the log-sum-exp normalizers of the two branches:
\begin{equation}
\alpha_i^{(l,h)}
=
\frac{\exp(\ell_{\text{sink},i}^{(l,h)})}
{\exp(\ell_{\text{sink},i}^{(l,h)}) +
\exp(\ell_{\text{comp},i}^{(l,h)} + \log g_{\text{comp}})},
\label{eq:method_alpha}
\end{equation}
\begin{equation}
O_i^{(l,h)}
=
\alpha_i^{(l,h)} O_{\text{sink},i}^{(l,h)}
+
\left(1-\alpha_i^{(l,h)}\right) O_{\text{comp},i}^{(l,h)}.
\label{eq:method_fused_output}
\end{equation}
The $\log g_{\text{comp}}$ term is a practical support-size correction for the pooled branch.
This row-wise weighting is useful because the benefit of compensation varies across query rows.
Some rows are already well explained by the sink support, whereas others benefit more from the pooled global summary.

\begin{algorithm}[t]
\caption{Online Sink-Sparse Hybrid Attention}
\label{alg:online_sink_hybrid}
\footnotesize
\begin{algorithmic}[1]
\Require queries $Q$, keys $K$, values $V$, static sink mask $M$, pooling gap $g_{\text{comp}}$
\Ensure attention output $O$
\Statex \textcolor{sinkteal}{\# static sparse sink branch}
\State $(O_{\text{sink}}, \ell_{\text{sink}}) \gets \operatorname{SparseAttn}(Q, K, V; M)$
\Statex \textcolor{sinkteal}{\# pooled-KV compensation branch}
\State $(K_{\text{pool}}, V_{\text{pool}}) \gets \operatorname{PoolKV}(K, V; g_{\text{comp}})$
\State $(O_{\text{comp}}, \ell_{\text{comp}}) \gets \operatorname{DenseAttn}(Q, K_{\text{pool}}, V_{\text{pool}})$
\Statex \textcolor{sinkteal}{\# normalization-aware adaptive fusion}
\For{each query row $i$}
    \State $\alpha_i \gets \frac{\exp(\ell_{\text{sink},i})}{\exp(\ell_{\text{sink},i}) + \exp(\ell_{\text{comp},i} + \log g_{\text{comp}})}$
    \State $O_i \gets \alpha_i O_{\text{sink},i} + (1-\alpha_i) O_{\text{comp},i}$
\EndFor
\State \Return $O$
\end{algorithmic}
\end{algorithm}

\subsection{Complexity and Deployment}

Let $\rho^{(l,h)}$ denote the token-level density induced by the static sink mask. Relative to dense per-head attention with dominant cost $\mathcal{O}(N^2)$, the hybrid operator can be summarized as
\begin{equation}
\begin{aligned}
\mathcal{C}_{\text{hyb}}^{(l,h)}
&= \mathcal{O}\!\left(\rho^{(l,h)}N^2 + \frac{N^2}{g_{\text{comp}}}\right) \\
&= \mathcal{O}\!\left(\left(\rho^{(l,h)} + g_{\text{comp}}^{-1}\right)N^2\right).
\end{aligned}
\label{eq:method_complexity}
\end{equation}
where the first term comes from the static sparse sink branch and the second term comes from the pooled-KV compensation branch, which keeps the query length at $N$ while reducing the key/value length to about $N / g_{\text{comp}}$. For reporting, we define the \emph{main-branch effective sparsity} as $1-\rho_{\text{main}}$, where $\rho_{\text{main}}$ denotes the runtime block-interaction density of the main sparse branch, i.e., the fraction of block interactions retained by its runtime block mask. The resulting design remains fully train-free and deployment-friendly: static sink masks are calibrated once for each backbone from a small dense run and then reused across future generations without retraining, fine-tuning, or modifying the denoising schedule.

\begin{table*}[t]
\centering
\caption{\textbf{Main T2V comparison on Wan2.1-T2V-1.3B under a unified evaluation setup.} \textbf{Dense} is the end-to-end reference under the same operator stack. Speedup is reported relative to this dense reference (\textbf{164.80s}). We report two representative operating points of \textbf{Ours-Hybrid}: a speed-oriented setting at $\tau=0.70$ and a quality-oriented setting at $\tau=0.85$. \textbf{\vbenchavg} is computed from the underlying full-precision per-dimension scores before rounding, so it may not exactly equal the mean of the displayed three-decimal values.}
\label{tab:t2v_main}
\setlength{\tabcolsep}{4pt}
\resizebox{\textwidth}{!}{%
\begin{tabular}{lccccccccccc}
\toprule
Method
& Time(s)$\downarrow$
& Speedup$\uparrow$
& SubCons.$\uparrow$
& MotSm.$\uparrow$
& DynDeg.$\uparrow$
& BgCons.$\uparrow$
& Scene$\uparrow$
& Aesth.$\uparrow$
& ImgQual$\uparrow$
& OverCons.$\uparrow$
& \vbenchavg$\uparrow$ \\
\midrule
Dense
& 164.80
& 1.00$\times$
& 0.894
& 0.972
& 0.861
& 0.941
& 0.145
& 0.588
& 0.614
& \textbf{0.235}
& 0.656 \\

SVG1~\cite{svg1}
& 115.49
& 1.43$\times$
& \textbf{0.941}
& 0.970
& 0.819
& 0.950
& 0.126
& 0.565
& 0.652
& 0.224
& 0.656 \\

SpargeAttn~\cite{zhang2025spargeattention}
& 139.68
& 1.18$\times$
& 0.930
& 0.954
& 0.833
& \textbf{0.961}
& \textbf{0.154}
& \textbf{0.608}
& \textbf{0.676}
& 0.208
& \textbf{0.665} \\
\midrule
Ours-Hybrid ($\tau=0.70$)
& \textbf{94.57}
& \textbf{1.74$\times$}
& 0.864
& 0.967
& 0.858
& 0.927
& 0.137
& 0.568
& 0.619
& 0.233
& 0.647 \\

Ours-Hybrid ($\tau=0.85$)
& 106.73
& 1.54$\times$
& 0.871
& \textbf{0.976}
& \textbf{0.931}
& 0.933
& 0.148
& 0.566
& 0.615
& 0.233
& 0.659 \\
\bottomrule
\end{tabular}%
}
\end{table*}

\section{Experiments}
\label{sec:exp}

We evaluate \textbf{Sink-Sparse Hybrid Attention} on both text-to-video and text-to-image generation. We use \textbf{Wan2.1-T2V-1.3B}~\cite{wan2025wan} as the T2V backbone and \textbf{Z-Image-Turbo}~\cite{zimage2025zimage} as the T2I backbone. Unless otherwise stated, each backbone uses its own model-specific static sink masks calibrated from a disjoint offline set.

\subsection{Experimental Setup}
\label{sec:exp_setup}

\paragraph{T2V setup.}
For text-to-video, we evaluate on \textbf{Wan2.1-T2V-1.3B} under a unified generation setting with resolution \textbf{832$\times$480}, \textbf{81 frames}, and \textbf{50 denoising steps}. For quality, we evaluate \textbf{eight VBench-1.0 dimensions}~\cite{huang2024vbench} and report their mean as \textbf{\vbenchavg}$\uparrow$.

\paragraph{T2I setup.}
For text-to-image, we evaluate on \textbf{Z-Image-Turbo}~\cite{zimage2025zimage} at \textbf{2048$\times$2048} resolution with \textbf{8 denoising steps}. We adopt the \textbf{GenEval2}~\cite{kamath2025geneval2} VQA evaluation protocol and report \textbf{\softtifagm}$\uparrow$ as the primary metric together with \textbf{\softtifaam}$\uparrow$ as an auxiliary metric.

\paragraph{Latency metric and hardware.}
For efficiency, we report end-to-end \textbf{Time(s)}$\downarrow$ and \textbf{Speedup}$\uparrow$. Unless otherwise stated, latency excludes model initialization and output writing. Time is measured on a single \textbf{NVIDIA A800-SXM4-80GB} GPU.

\paragraph{Calibration and implementation details.}
Following Sec.~\ref{sec:method}, \textbf{Ours-Sink} denotes the sink-preserving sparse branch alone, while \textbf{Ours-Hybrid} denotes the full method with the pooled-KV compensation branch and adaptive fusion. Unless otherwise stated, both variants use static sink masks identified from calibration runs over \textbf{4 prompts} and \textbf{2 noise seeds}, with attention statistics aggregated across denoising steps. The calibration set is disjoint from the downstream evaluation set. We use a fixed consecutive block partition for mask acquisition, and set $g_{\text{comp}}=30$ for \textbf{Ours-Hybrid}. For Z-Image-Turbo, we keep the same overall pipeline and calibrate a separate mask set for the image backbone.

\begin{table}[t]
\centering
\caption{\textbf{Main T2I comparison on Z-Image-Turbo.} Evaluated at $2048\times2048$ resolution with 8 denoising steps. We report \textbf{\softtifaam}, \textbf{\softtifagm}, end-to-end \textbf{Time(s)}, and \textbf{Speedup} relative to the dense baseline.}
\label{tab:t2i_main}
\setlength{\tabcolsep}{6pt}
\resizebox{\columnwidth}{!}{%
\begin{tabular}{lcccc}
\toprule
Method & \softtifaam$\uparrow$ & \softtifagm$\uparrow$ & Time(s)$\downarrow$ & Speedup$\uparrow$ \\
\midrule
Dense & \textbf{67.80} & \textbf{16.38} & 23.39 & 1.00$\times$ \\
SpargeAttn~\cite{zhang2025spargeattention} & 66.88 & 15.07 & 15.92 & 1.47$\times$ \\
Ours-Hybrid ($\tau=0.85$) & 66.48 & 15.21 & \textbf{14.88} & \textbf{1.57$\times$} \\
\bottomrule
\end{tabular}%
}
\end{table}

\paragraph{Unified operator stack and baselines.}
To ensure a fair, implementation-consistent, and runtime-comparable comparison throughout evaluation, for \textbf{Dense}, exact full self-attention is implemented with \textbf{FlashAttention}~\cite{dao2022flashattention}. For the T2V comparison, \textbf{Dense}, \textbf{SVG1}~\cite{svg1}, \textbf{SpargeAttn}~\cite{zhang2025spargeattention}, and our method also share the same optimized implementations for other common attention components, namely \textbf{QK-norm} and \textbf{RoPE}, adopted directly from the SVG1 codebase for consistency. For T2V, we compare against \textbf{Dense}, \textbf{SVG1}~\cite{svg1}, and \textbf{SpargeAttn}~\cite{zhang2025spargeattention}. For T2I, we compare against \textbf{Dense} and \textbf{SpargeAttn}.

\begin{figure}[t]
    \centering

    {\footnotesize \textbf{Wan2.1-T2V-1.3B}}\par
    \textbf{Dense}\par
    \safeincludegraphics[width=0.99\linewidth]{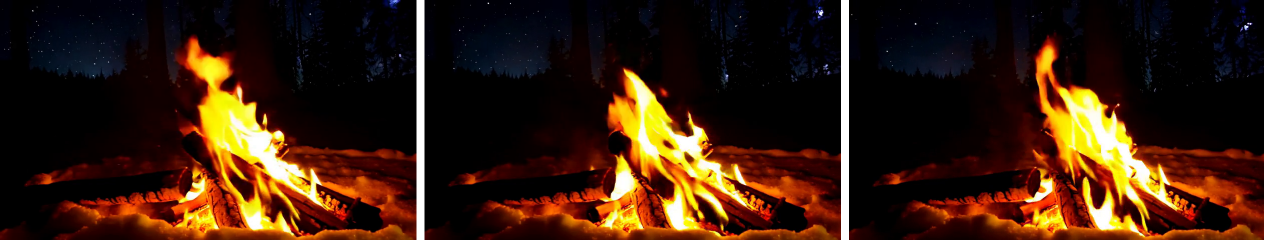}

    \textbf{Ours-Hybrid ($\tau=0.85$)}\par
    \safeincludegraphics[width=0.99\linewidth]{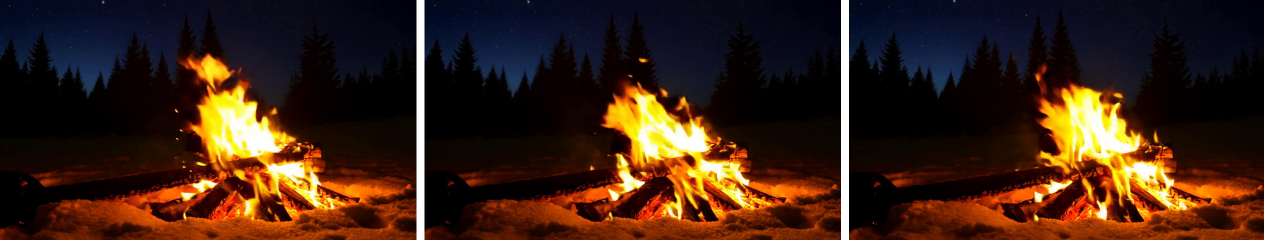}

    \textbf{SpargeAttn}\par
    \safeincludegraphics[width=0.99\linewidth]{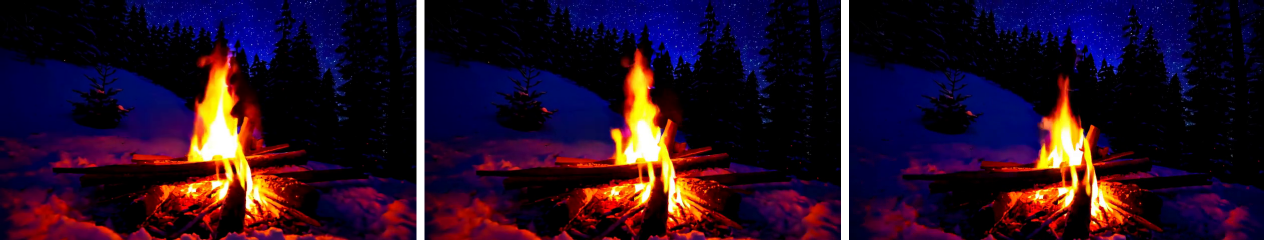}

    \textbf{SVG1}\par
    \safeincludegraphics[width=0.99\linewidth]{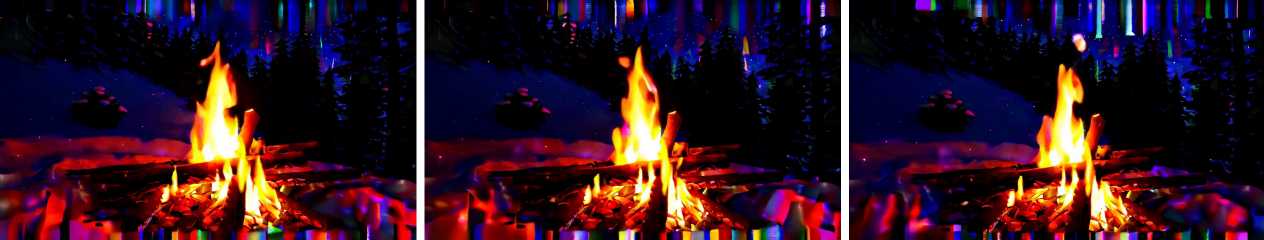}

    \vspace{3pt}
    {\footnotesize \textbf{Z-Image-Turbo}}\par
    \begin{minipage}{0.99\linewidth}
        \centering
        {\footnotesize
        \makebox[\dimexpr\linewidth/3\relax][c]{Dense}%
        \makebox[\dimexpr\linewidth/3\relax][c]{Ours-Hybrid ($\tau=0.85$)}%
        \makebox[\dimexpr\linewidth/3\relax][c]{SpargeAttn}\par
        }
        \vspace{2pt}
        \safeincludegraphics[width=\linewidth]{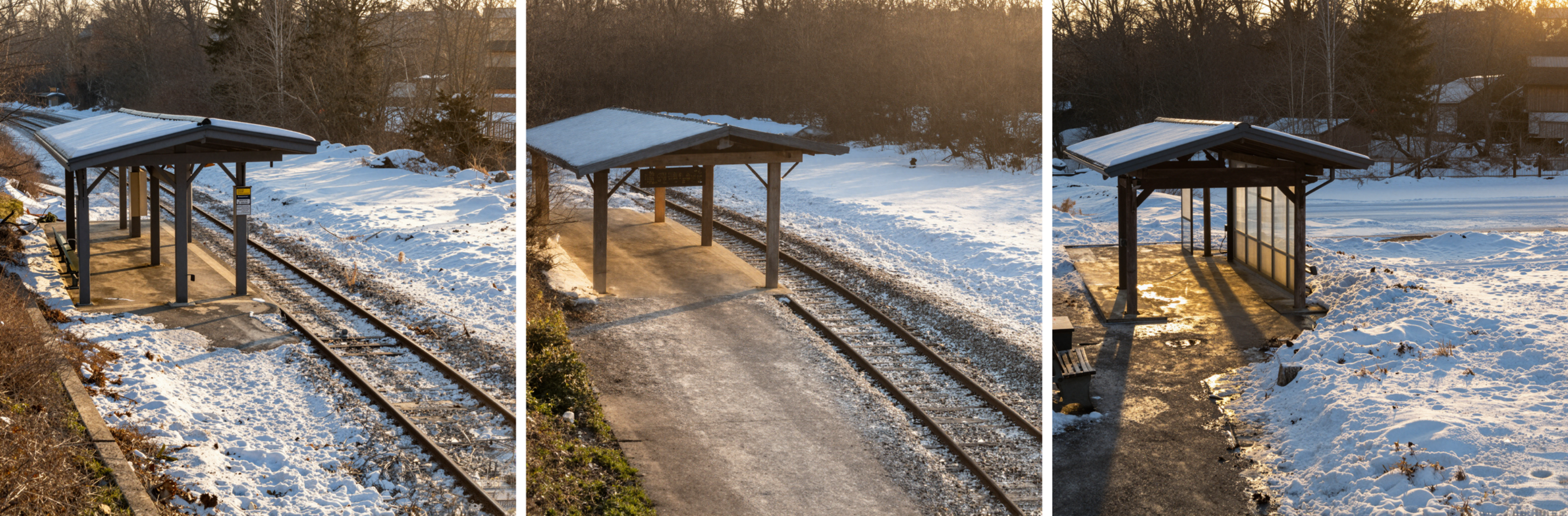}
    \end{minipage}

    \caption{\textbf{Representative qualitative comparison on Wan2.1-T2V-1.3B and Z-Image-Turbo.} For Wan2.1-T2V-1.3B, we show three synchronized frames (frames 16, 40, and 64 out of 81) for the prompt ``Campfire at night in a snowy forest with starry sky in the background.'' using the same noise seed across methods. For Z-Image-Turbo, we show image results for the prompt ``Winter station at golden hour.'' in the order Dense, Ours-Hybrid ($\tau=0.85$), and SpargeAttn.}
    \Description{Single-column qualitative comparison on both backbones. The upper part shows Wan2.1-T2V-1.3B outputs for the prompt 'Campfire at night in a snowy forest with starry sky in the background.' with four rows: Dense, Ours-Hybrid with tau equals 0.85, SpargeAttn, and SVG1; each row contains three synchronized frames sampled at frames 16, 40, and 64 out of 81. The lower part shows Z-Image-Turbo outputs for the prompt 'Winter station at golden hour.' in the order Dense, Ours-Hybrid with tau equals 0.85, and SpargeAttn.}
    \label{fig:qual_video_main}
\end{figure}

\subsection{Main Results}
\label{sec:exp_main}

Table~\ref{tab:t2v_main} reports the main T2V comparison. \textbf{Ours-Hybrid} exposes two useful operating points. At $\tau=0.70$, it is the fastest setting in the table, reducing end-to-end time from \textbf{164.80s} to \textbf{94.57s} (\textbf{1.74$\times$}) with \textbf{0.647} \textbf{\vbenchavg}. At $\tau=0.85$, it reaches \textbf{1.54$\times$} speedup while improving \textbf{\vbenchavg} to \textbf{0.659}. This quality-oriented setting is slightly better than \textbf{SVG1} in overall average (\textbf{0.659} vs.\ \textbf{0.656}) while remaining faster (\textbf{106.73s} vs.\ \textbf{115.49s}). Compared with \textbf{SpargeAttn}, it gives up a small amount of average quality (\textbf{0.659} vs.\ \textbf{0.665}) in exchange for a much larger runtime reduction (\textbf{1.54$\times$} vs.\ \textbf{1.18$\times$}). This indicates that hybrid compensation is useful because it retains most of the speed benefit of static sparsity while making the quality-oriented point much more competitive.

Figure~\ref{fig:qual_video_main} provides representative qualitative examples on both backbones. Relative to the sparse baselines, \textbf{Ours-Hybrid} remains visually closer to the dense reference in overall structure and appearance, which is consistent with the quantitative trade-offs in Tables~\ref{tab:t2v_main} and~\ref{tab:t2i_main}.

Table~\ref{tab:t2i_main} reports the T2I comparison. \textbf{Ours-Hybrid} at $\tau=0.85$ reduces end-to-end runtime from \textbf{23.39s} to \textbf{14.88s}, yielding \textbf{1.57$\times$} speedup. Under the GenEval2 VQA protocol, it attains \textbf{15.21} \softtifagm{} and \textbf{66.48} \softtifaam{}. Compared with \textbf{SpargeAttn}, it is faster (\textbf{1.57$\times$} vs.\ \textbf{1.47$\times$}) and slightly better on the primary \softtifagm{} metric (\textbf{15.21} vs.\ \textbf{15.07}), while remaining close on \softtifaam{} (\textbf{66.48} vs.\ \textbf{66.88}). The T2I result also suggests that the hybrid branch is a general global-context complement rather than a video-specific heuristic. These results indicate that the same train-free framework transfers to the two tested backbones.

\subsection{Efficiency--Quality Trade-off}
\label{sec:exp_efficiency}

\begin{figure*}[t]
    \centering
    \safeincludegraphics[width=0.90\textwidth]{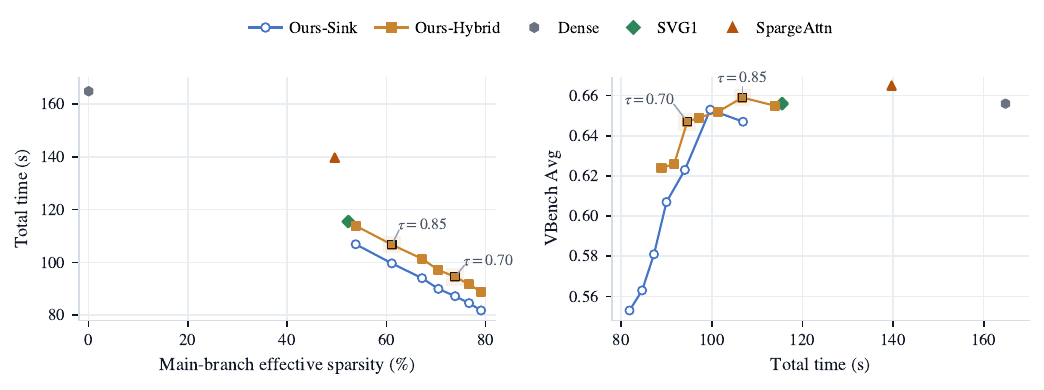}
    \caption{\textbf{Efficiency profile and quality--efficiency frontier in the T2V setting.} \textbf{Left:} End-to-end runtime versus main-branch effective sparsity for \textbf{Ours-Sink} and \textbf{Ours-Hybrid}, with \textbf{Dense}, \textbf{SVG1}, and \textbf{SpargeAttn} shown as reference points. \textbf{Right:} Quality--efficiency frontier measured by \textbf{\vbenchavg} versus end-to-end runtime. Lower coverage thresholds yield stronger acceleration, while higher thresholds improve quality; $\tau=0.70$ and $\tau=0.85$ are highlighted as representative speed-oriented and quality-oriented operating points, respectively.}
    \Description{A two-panel figure summarizing the efficiency and trade-off behavior of the proposed method in the T2V setting. The left panel plots end-to-end runtime against main-branch effective sparsity for Ours-Sink and Ours-Hybrid, together with Dense, SVG1, and SpargeAttn as reference points. The right panel plots VBench Avg against end-to-end runtime and highlights tau equals 0.70 and tau equals 0.85 as representative operating points.}
    \label{fig:eff_tradeoff}
\end{figure*}

Figure~\ref{fig:eff_tradeoff} summarizes the efficiency trend and the corresponding quality--efficiency frontier in the T2V setting. The left panel shows that lower coverage thresholds produce sparser main branches and lower runtime for both \textbf{Ours-Sink} and \textbf{Ours-Hybrid}. For example, \textbf{Ours-Sink} at $\tau=0.60$ reaches \textbf{79.09\%} main-branch effective sparsity and reduces end-to-end time to \textbf{81.82s} (\textbf{2.01$\times$}), while the pooled-KV compensation branch adds a modest latency overhead and recovers a substantial part of the quality gap. At comparable sparsity budgets, \textbf{Ours-Hybrid} consistently shifts the operating frontier upward relative to \textbf{Ours-Sink}, which further confirms that the hybrid attention is useful rather than redundant: the extra branch converts a small amount of additional computation into a noticeably better quality--efficiency trade-off.

The right panel shows the corresponding quality--efficiency frontier for \textbf{Ours-Hybrid}. Increasing $\tau$ improves quality up to a peak at \textbf{$\tau=0.85$}, whereas \textbf{$\tau=0.70$} provides the stronger speed-oriented operating point. The frontier also shows a clear knee between $\tau=0.70$ and $\tau=0.85$: below this range, further sparsification yields additional speed but a more visible quality drop; above it, the masks become denser yet the incremental quality gain is smaller.

\subsection{Calibration Size}
\label{sec:abl_calib_size}

Because the static sink masks are identified offline, an important practical question is how much calibration data is needed. We study this effect on \textbf{Wan2.1-T2V-1.3B} for \textbf{Ours-Hybrid} at the representative speed-oriented setting \textbf{$\tau=0.70$}, while keeping all inference hyperparameters fixed. A calibration trajectory denotes one prompt--seed pair whose attention statistics are aggregated over the full 50-step denoising process. We evaluate four calibration budgets, namely \textbf{2$\times$2}, \textbf{4$\times$2}, \textbf{4$\times$4}, and \textbf{8$\times$4} prompt--seed configurations. All calibration prompts are disjoint from the downstream evaluation set.

To keep the presentation compact, we report only end-to-end \textbf{Time(s)}$\downarrow$, \textbf{\vbenchavg}$\uparrow$, and the one-time \textbf{Offline calibration cost}$\downarrow$. The offline cost is estimated from the shared-stack dense runtime (\textbf{164.80s} per prompt--seed trajectory), excluding model initialization and video writing, and is reported in minutes as an approximate deployment-oriented reference.

\begin{table}[t]
\centering
\caption{\textbf{Effect of calibration set size for Ours-Hybrid at $\tau=0.70$ on Wan2.1-T2V-1.3B.} Calibration budget is specified by the number of prompts and seeds used for offline mask acquisition, where each prompt--seed pair contributes one full denoising trajectory. \textbf{Offline calibration cost} is a rough one-time estimate computed from the shared-stack dense latency (\textbf{164.80s} per trajectory), reported in minutes. Larger calibration budgets improve quality modestly, and the gains largely saturate by \textbf{4$\times$4}.}
\label{tab:abl_calib_size}
\setlength{\tabcolsep}{5pt}
\resizebox{\columnwidth}{!}{%
\begin{tabular}{lccc}
\toprule
Calibration Budget & Time(s)$\downarrow$ & Offline Calib. Cost$\downarrow$ & \vbenchavg$\uparrow$ \\
\midrule
$2\times 2$ & 92.84 & 11.0 min & 0.645 \\
$4\times 2$ & 94.57 & 22.0 min & 0.647 \\
$4\times 4$ & 93.71 & 43.9 min & 0.650 \\
$8\times 4$ & 94.18 & 87.9 min & 0.649 \\
\bottomrule
\end{tabular}%
}
\end{table}

Table~\ref{tab:abl_calib_size} shows that enlarging the calibration budget brings modest but nonzero quality gains. The most visible gain appears when increasing the budget from \textbf{2$\times$2} to the default \textbf{4$\times$2} setting, where \textbf{\vbenchavg} improves from \textbf{0.645} to \textbf{0.647} with only a small runtime difference (\textbf{92.84s} to \textbf{94.57s}). Increasing the calibration size to \textbf{4$\times$4} further improves \textbf{\vbenchavg} to \textbf{0.650}. The largest budget, \textbf{8$\times$4}, yields \textbf{0.649}, which is effectively on par with \textbf{4$\times$4} rather than clearly better. This indicates that the quality gains largely saturate once the calibration set becomes moderately large.

Two additional observations are worth noting. First, runtime stays nearly unchanged across calibration budgets, which means that calibration mainly affects \emph{which} block interactions are retained rather than \emph{how many} interactions are executed at test time. Second, the saturation around \textbf{4$\times$4} suggests that the dominant layer/head-specific sink structure becomes stable after a moderate offline budget, making \textbf{4$\times$2} or \textbf{4$\times$4} a practical trade-off between one-time calibration cost and downstream quality.

\subsection{Effect of Hybrid Compensation}
\label{sec:abl_comp}

Table~\ref{tab:abl_comp} shows that, on \textbf{Wan2.1-T2V-1.3B}, the pooled-KV compensation branch consistently improves \textbf{\vbenchavg} over the corresponding \textbf{Ours-Sink} setting across all tested thresholds. The gain is largest in the more aggressive regime, reaching \textbf{+0.071} at $\tau=0.60$, \textbf{+0.063} at $\tau=0.65$, and \textbf{+0.066} at $\tau=0.70$. As $\tau$ increases and the static sink mask becomes denser, the quality gap narrows, but the hybrid branch still remains beneficial.

The trend is informative. The compensation branch contributes most when the main sparse branch is aggressive and therefore more likely to miss weaker or more distributed context. Once $\tau$ becomes large and the sink branch already preserves most dominant support, the omitted context shrinks and the additional gain becomes marginal. This behavior is desirable for deployment: the extra branch pays for itself primarily in the operating regime where static sparsification alone would otherwise lose the most quality.

\begin{table}[t]
\centering
\caption{\textbf{Effect of the pooled-KV compensation branch across coverage thresholds on Wan2.1-T2V-1.3B.} \textbf{Ours-Sink} uses only the static sink mask, while \textbf{Ours-Hybrid} additionally enables the pooled-KV compensation branch with adaptive fusion. $\Delta$Avg denotes the gain of \textbf{Ours-Hybrid} over \textbf{Ours-Sink}.}
\label{tab:abl_comp}
\setlength{\tabcolsep}{4pt}
\resizebox{\columnwidth}{!}{%
\begin{tabular}{cccccc}
\toprule
$\tau$
& Sink Time$\downarrow$
& Sink Avg$\uparrow$
& Hybrid Time$\downarrow$
& Hybrid Avg$\uparrow$
& $\Delta$Avg$\uparrow$ \\
\midrule
0.60 & 81.82 & 0.553 & 88.88 & 0.624 & +0.071 \\
0.65 & 84.60 & 0.563 & 91.69 & 0.626 & +0.063 \\
0.70 & 87.22 & 0.581 & 94.57 & 0.647 & +0.066 \\
0.75 & 89.98 & 0.607 & 97.12 & 0.649 & +0.042 \\
0.80 & 94.04 & 0.623 & 101.37 & 0.652 & +0.029 \\
0.85 & 99.64 & 0.653 & 106.73 & 0.659 & +0.006 \\
0.90 & 106.88 & 0.647 & 113.89 & 0.655 & +0.008 \\
\bottomrule
\end{tabular}
}
\end{table}

\section{Conclusion}
\label{sec:conclusion}

We presented \textbf{Sink-Sparse Hybrid Attention}, a train-free framework that identifies reusable layer/head-specific sink structure from a small offline calibration set, preserves it with a static sparse branch, and complements it with a lightweight pooled-KV branch. Experiments on T2V and T2I backbones demonstrate favorable quality--efficiency trade-offs, achieving \textbf{1.54$\times$} end-to-end speedup on \textbf{Wan2.1-T2V-1.3B} and \textbf{1.57$\times$} on \textbf{Z-Image-Turbo}. The results clarify why the hybrid approach is practically useful: it restores coarse omitted context at modest cost, making aggressive train-free sparsification more robust than a sink-only design. These results suggest that reusable importance-aware sparse support is a practical acceleration path for pretrained DiTs.

\paragraph{Limitations.}
The current evaluation covers a limited set of backbones, and the method requires backbone-specific offline calibration. Additionally, fixed static masks may adapt less well to distribution shifts or unusual prompts whose salient interactions differ from the calibration set. The current study also does not examine whether masks transfer reliably across backbones. Under aggressive sparsity, rare or strongly input-dependent dependencies may still be lost.

\bibliographystyle{ACM-Reference-Format}
\bibliography{main}

\end{document}